\documentclass{article}

\usepackage{times}
\usepackage{graphicx} 
\usepackage{subfigure}

\usepackage{natbib}

\usepackage{algorithm}
\usepackage{algorithmic}

\usepackage{hyperref}

\usepackage{tikz}
\usetikzlibrary{matrix,shapes,arrows,positioning,chains,calc,shapes.geometric}
\usepackage{adjustbox}
\usepackage{courier}
\usepackage{paralist}
\newcommand{\m}{\ensuremath}

\tikzstyle{startstop} = [rectangle, rounded corners, minimum width=3cm, minimum height=1cm,text centered, draw=black, fill=red!30]
\tikzstyle{io} = [rectangle, minimum width=2cm, minimum height=1cm, text centered, text width=3cm,draw=black, fill=blue!30]
\tikzstyle{process} = [rectangle, minimum width=3cm, minimum height=1cm, text centered, text width=3cm, draw=black, fill=orange!30]
\tikzstyle{decision} = [diamond, minimum width=3cm, minimum height=1cm, text centered, text width=3cm, draw=black, fill=green!30]
\tikzstyle{arrow} = [thick,->,>=stealth]

\usepackage[accepted]{data4good2016}

\icmltitlerunning{Machine Learning for Antimicrobial Resistance}
\begin{document}

\twocolumn[
\icmltitle{Machine Learning for Antimicrobial Resistance}

\icmlauthor{John W. Santerre}{santerre@cs.uchicago.edu}
\icmladdress{Department of Computer Science, University of Chicago, Chicago, IL USA}
\vspace{-4pt}
\icmlauthor{James J. Davis}{jjdavis@anl.gov}
\icmladdress{Argonne National Laboratory,
      Argonne, IL USA}
\vspace{-4pt}
\icmlauthor{Fangfang Xia}{fangfang@anl.gov}
\icmladdress{Argonne National Laboratory,
      Argonne, IL USA}
\vspace{-4pt}
\icmlauthor{Rick Stevens}{stevens@anl.gov}
\icmladdress{Argonne National Laboratory,
      Argonne, IL USA}
\vspace{-4pt}
\icmlkeywords{applied machine learning, AMR, antibiotic resistance, random forest, computational biology, machine learning, ICML}

\vskip 0.3in
]
\vspace{-10pt}
\begin{abstract}
Biological datasets amenable to applied machine learning are more available today than ever before, yet they lack adequate representation in the Data-for-Good community. Here we present a work in progress case study performing analysis on antimicrobial resistance (AMR) using standard ensemble machine learning techniques and note the successes and pitfalls such work entails. Broadly, applied machine learning (AML) techniques are well suited to AMR, with classification accuracies ranging from mid-\m{90 \%} to low-\m{80\%} depending on sample size. Additionally, these techniques prove successful at identifying gene regions known to be associated with the AMR phenotype. We believe that the extensive amount of biological data available, the plethora of problems presented, and the global impact of such work merits the consideration of the Data-for-Good community.\vspace{-10pt}
\end{abstract}
\vspace{-10pt}
\section{Introduction}
Applied data analysis has seen explosive growth in popular interest in recent years. In parallel with this growth, there has been an increased interest in ``democratizing'' the tools and techniques for use by communities who might not have access otherwise. In many ways the machine learning problems faced by such communities are identical to the needs of larger institutions (i.e. return on investment optimization is technically similar regardless of the tax status of the company). Yet, significant shortage of skilled workers has created salary requirements that prevent nonprofits from competing. This reflects a broader trend in the nonprofit sector struggling to find IT talent. Exacerbating this problem are the struggles associated with trying to integrate a machine learning expert/team into an existing business structure or software stack. To counteract this trend organizations for several years have made great strides both in training new individuals perform data analysis and encouraging advanced practitioners (e.g. PhD graduates in Computer Science or Statistics) to pursue this work. The Data-for-Good community (such as Data Science for Social Good fellowship and the IBM Social Good Fellowship) has achieved remarkable success in promoting work directly with not-for-profit organizations, non-government organizations, or local governments. Here we are encouraging machine learning practitioners to consider the analysis of public biological datasets with a focus on problems that would have the greatest value to medical facilities in underserved communities. Using our work on genotype to phenotype classification of AMR as a case study, we note how AML techniques can yield real advances in the field. Biology and medicine are rich with a multitude of problems, many of which are very niche big business investments. We hope to attract the attention of the Data-for-Good community to this class of problems which we believe will be an area of growth, both in and out of the Data-for-Good community, over the next decade. Here we present work in progress, which was originally intended to demonstrate to biologists the power AML can have on many problems they are interested in. We believe it also serves as a good case study for the Data-for-Good movement: our use of off-the-shelf classifiers applied to public datasets to answer unique real-world problems shares a great similarity with the type of problems the Data-for-Good community attempts to solve.
\vspace{-10pt}
\section{Background}
The very foundation of computational biology rests on the application of computer science to biological problems. Similarly, the biostatistics community has a long history of investigating the unique statistical problems that biology presents. Well-funded Silicon Valley startups, such as 23andMe, are in the business of collecting large amounts of data and applying statistical techniques typical of applied machine learning or ``big data'' analysis. Meanwhile, Deep Genomics targets particularly tricky biological problems with both copious data and advanced deep learning techniques. Historically, advances in computational biology have been achieved with well targeted small datasets or datasets which are hard to generate. In contrast, current growth in the field is typified by large scale aggregation of data using various protocols, standards and databases. We believe this aggregation of large datasets will create a unique opportunities for a variety AML solutions and most importantly increased involvement of practitioners.
\vspace{-10pt}
\subsection{Data-for-Good Aspects of AMR}
AMR is a well recognized problem with profound implications for global health \cite{centres2013antibiotic,world2014antimicrobial,house2015national}. In many biological disciplines grant application guidelines or funding agencies typically require that the data generated and used for analysis in the course of a study are made publicly available. 
This allows unique opportunity for large-scale aggregation platforms, such as Pathosystems Resource Integration Center (PATRIC) \cite{patric} to serve a unique role. These platforms provide greater access to data but are currently geared toward professional biologists. Some of their pros/cons are:

\begin{description}
  \vspace{-12pt}
  \item \textbf{Advantages}
  \vspace{-2pt}
\begin{itemize}
\item Global impact
\item Rich datasets
\item Ample availability of access
\item Presents applied and theoretical problems
\end{itemize}
\vspace{-10pt}
  \item \textbf{Disadvantages}
  \vspace{-2pt}
\begin{itemize}
\item Minimal documentation
\item Experts required
\item Easy to misinterpret
\item Unclear how to operationalize analytical results
\vspace{-6pt}
\end{itemize}
\end{description}

We note that the principal disadvantages listed above are mainly related to putting the data into a format that is easily accesible to an AML professional, rather than a professional biologist. We note that these problems are not unique to computational biology and have been surmounted in other domains either through standardization of the data processing or attachment of metadata.

\begin{figure}[!ht]
\centering
\includegraphics[width=0.4 \textwidth]{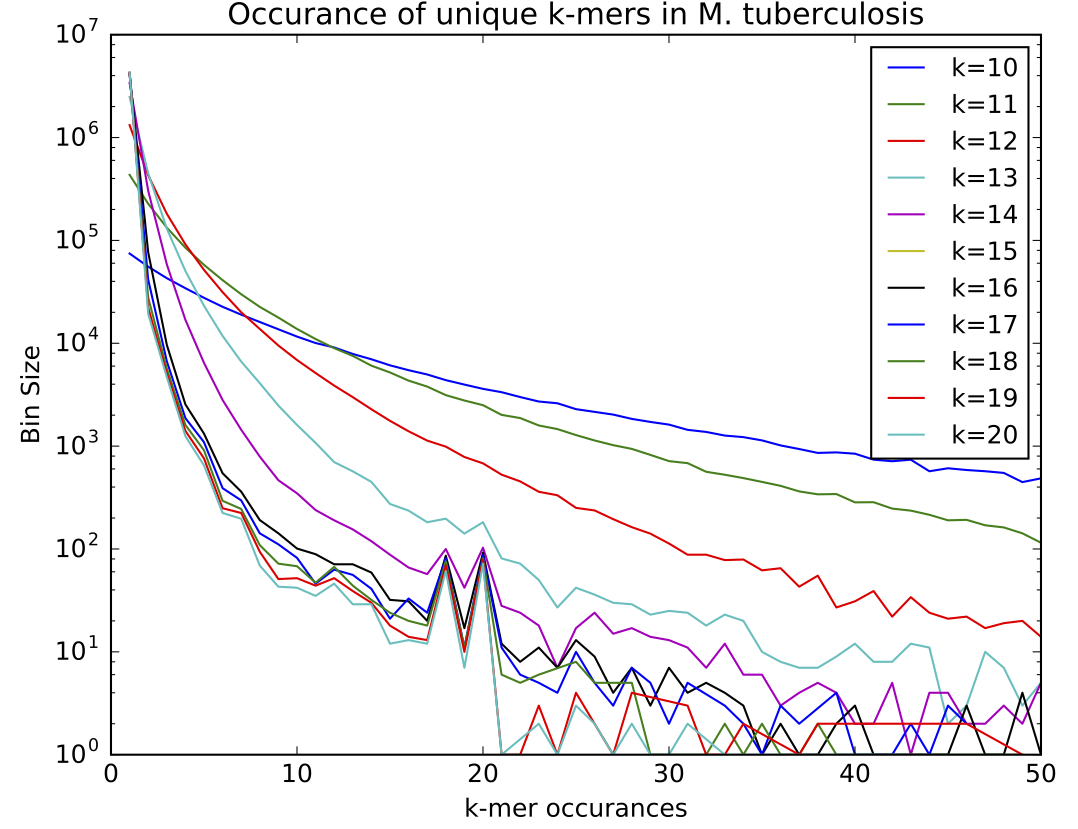}
\caption{Occurrence of unique \m{k}-mers in the MTB strain 1773.372. \m{y}-axis shows how many times a \m{k}-mer count (\m{x}-axis) appeared.\label{fig:MTB2}}
\vspace{-6pt}
\end{figure}

\begin{figure}[!ht]
\centering
\includegraphics[width=0.4 \textwidth]{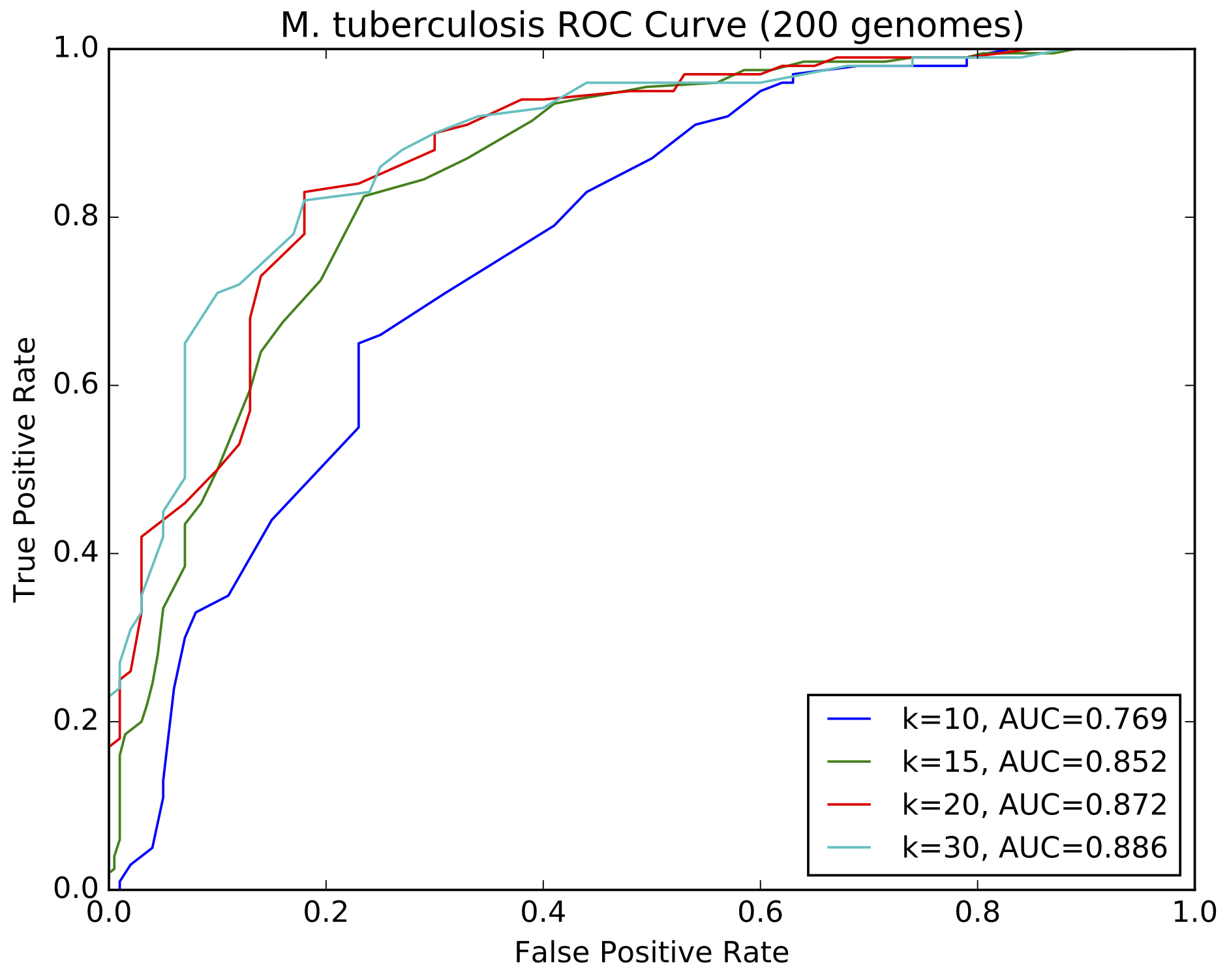}
\caption{Adabost performed on MTB strain Rifampicin resistant\label{fig:5}}
\vspace{-12pt}
\end{figure}

\begin{table}[t]
\caption{A comparison of the relationship between choice of \m{k}-mer length, the total number of unique \m{k}-mers, and the size of the dataset on disk.}
\label{tab:kmer-size-table}
\vskip 0.15in
\begin{center}
\begin{small}
\begin{sc}
\begin{tabular}{lrlr}
\hline
\abovespace\belowspace
\m{k} & \m{k}-mers counts & size \\
\hline
\abovespace
10  & 518304   & 214MB \\
11  & 1,857,213  &  733MB \\
12  & 5,058,522  &  2.0GB\\
13  & 9,502,404  &  3.7GB\\
14  & 13,171,189 &  5.1GB\\
15  & 15,258,381 &  6.0GB\\
20  & 17.653,652 &  7.0GB\\
25  & 18,816,555 &  7.5GB\\
30  & 19,882,628 &  8.0GB\\
\hline
\end{tabular}
\end{sc}
\end{small}
\end{center}
\vspace{-12pt}
\end{table}
\vspace{-10pt}
\section{AMR Case Study}
Identification of regions and genes associated with antimicrobial resistance is an extensively studied topic. While we have analysed a variety of datasets (Table \ref{tab:datasetsTable}), here we focus on principally three different species of bacteria--\textit{Streptococcus pneumoniae} ($\beta$-lactams resistant), \textit{Acinetobacter baumannii} (carbapenems resistant) and \textit{Mycobacterium tuberculosis} (isoniazid resistant) \cite{chewapreecha_dense_2014}.

\begin{table}[t]
\caption{Sample set of currently available datasets on the Pathosystems Resource Integration Center (PATRIC) system. SUS and RES designate susceptible and resistant strains, respectively.}
\label{tab:datasetsTable}
\vskip 0.15in
\begin{center}
\begin{small}
\begin{sc}
\begin{tabular}{llcc}
\hline
\abovespace\belowspace
\textbf{Strain name} & \textbf{Antibiotic} &\textbf{SUS}& \textbf{RES} \\
\hline
\abovespace
\textit{A. baumannii }& carbapenem & 110 & 122 \\
\textit{M. tuberculosis} & ethambutol & 878 & 464 \\
\textit{M. tuberculosis} & ethionamide & 283 & 174 \\
\textit{M. tuberculosis} & isoniazid & 580 & 986 \\
\textit{M. tuberculosis} & kanamycin & 504 & 202 \\
\textit{M. tuberculosis} & moxifloxacin & 277 & 103 \\
\textit{M. tuberculosis} & ofloxacin & 492 & 267 \\
\textit{M. tuberculosis} & rifampicin & 545 & 715 \\
\textit{M. tuberculosis} & streptomycin & 883 & 571 \\
\textit{S. aureus }& methicillin & 115 & 491 \\
\textit{S. pneumoniae} & beta-lactam & 1505 & 1563 \\
\textit{S. pneumoniae}& trimoxazole & 585 & 2126 \\

\hline
\end{tabular}
\end{sc}
\end{small}
\end{center}
\vskip-0.1in
\end{table}

\begin{figure}[!ht]
\centering
\includegraphics[width=0.45 \textwidth]{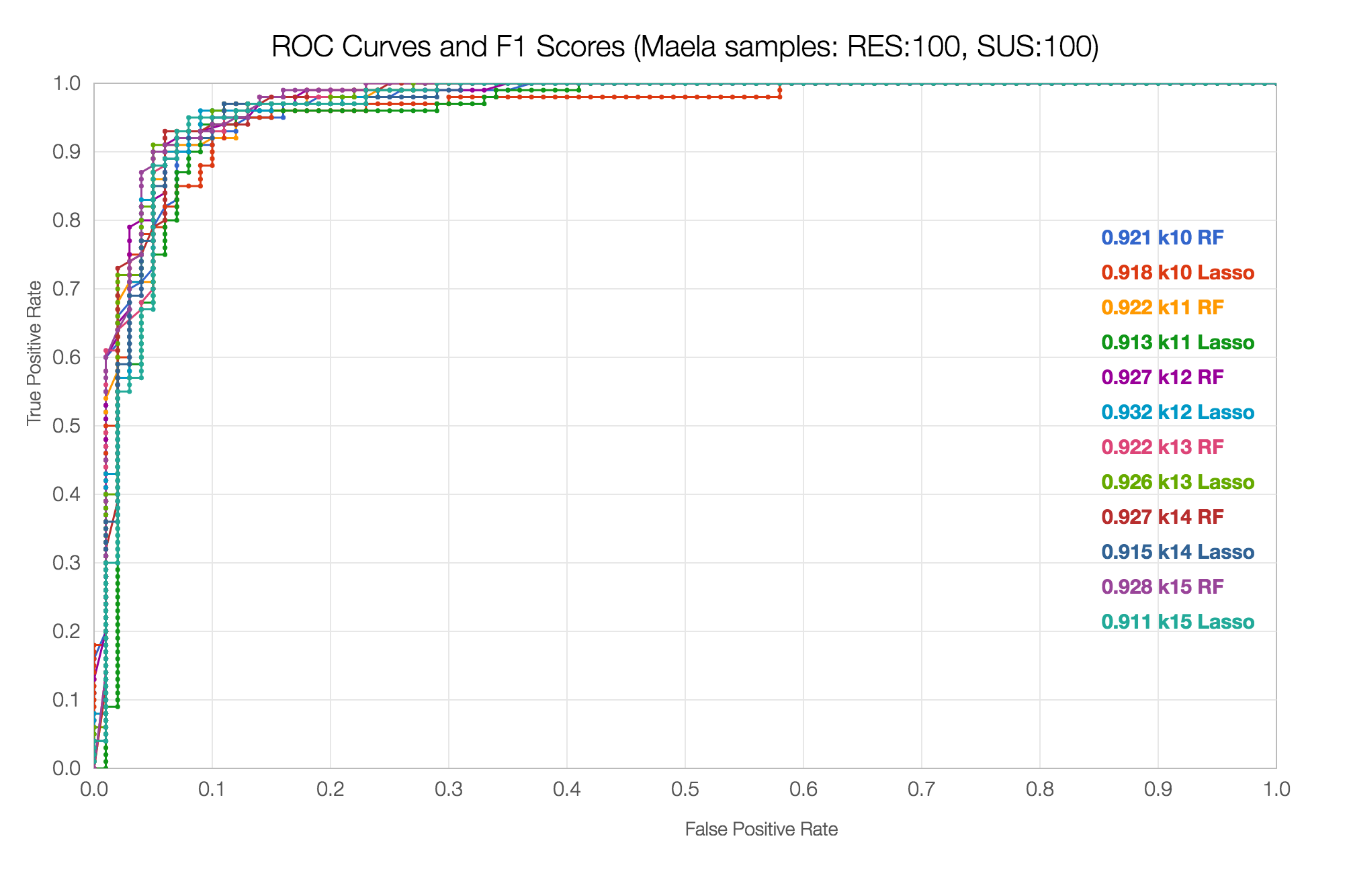}
\caption{ROC plots for RF and Lasso. RF is trained on \m{100} trees.\label{fig:1}}
\vspace{-12pt}
\end{figure}

\subsection{Analysis}
The Random Forest (RF) analysis \cite{breiman_random_2001} was performed using the default settings of the \texttt{scikit-learn} API \cite{pedregosa_scikit-learn:_2011}. Other algorithms (e.g. Adaboost \cite{freund1997decision}, logistic regression and deep learning among others) were investigated, however these results are largely beyond the scope of this paper. All analysis was performed using Python 2.7 on a 32 core machine with 1TB RAM.
\vspace{-10pt}
\subsection{Data Preprocessing}
Assembled DNA contigs (partially assembled medium length strands of DNA that represent each isolate) were converted to \m{k}-mers (fixed length short strands typically of length of \m{10-50}). Each isolate is represented by its unique \m{k}-mer count as features. Table \ref{tab:kmer-size-table} shows how the choice of \m{k} effects \m{k}-mer count and final dataset size. We have two variables we can tune--the first is the size of \m{k}, the second is the number of isolates to consider. Setting \m{k=1} is equivalent to finding the cellular GC-content (i.e. the percent of guanine (G) or cytosine (C) nitrogenous bases on the DNA molecule), which is used for coarse-grained analysis in computational biology. We note that overall size of the dataset matrix expands rapidly at roughly around \m{k=13} (Figure \ref{fig:MTB2}). In addition, as demonstrated in Figure \ref{fig:5}, a larger \m{k} yields a better ROC metric. Conversely, on another dataset (Figure \ref{fig:1}) we note that both RF and Lasso perform fairly similarly regardless of the size of \m{k}. We believe further investigation of the generalizability across species and phenotypes is necessary and have noted when our analysis included \m{k}-mers of different lengths.

\begin{figure}[!ht]
\centering
\includegraphics[width=0.45 \textwidth]{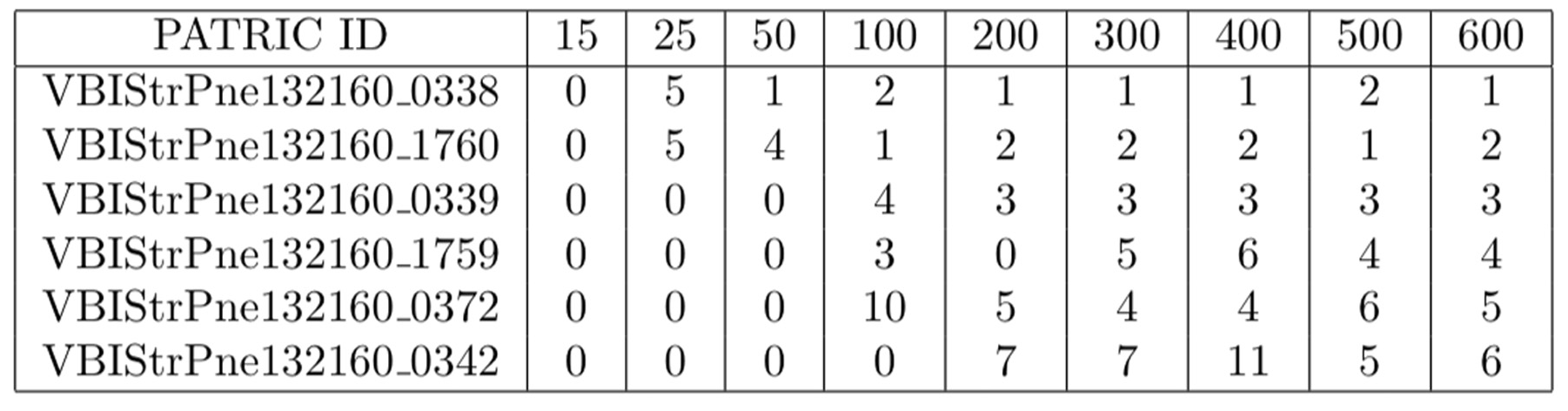}
\caption{A list of the top gene regions, by PATRIC ID, identified from feature importance calculated using RF on decreasing sample sizes of MTB.  Note that ultimately the top gene region places in top \m{5} with as little as \m{25} samples.
\label{fig:figure3}}
\vspace{-12pt}
\end{figure}

\subsection{Overview of Results}
\begin{figure}[!ht]
\centering
\centerline{ \small{\m{\beta}-lactam resistant \textit{S. pneumoniae}}}
\includegraphics[width=0.37\textwidth]{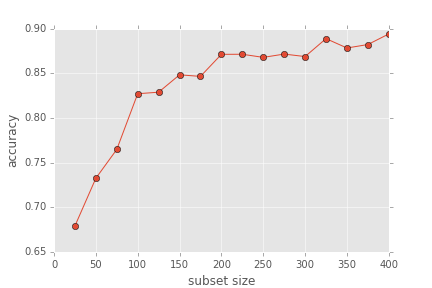}\vspace{9pt}

\centerline{\small{Carbapenem resistant \textit{A. baumannii}}}
\includegraphics[width=0.37\textwidth]{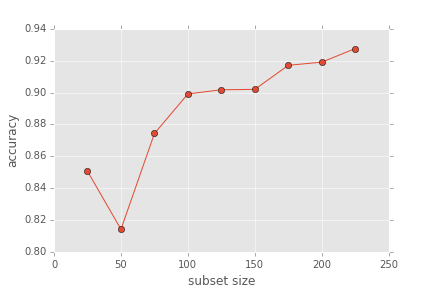}

\centerline{\small{Isoniazid resistant \textit{M. tuberculosis}}}
\includegraphics[width=0.37\textwidth]{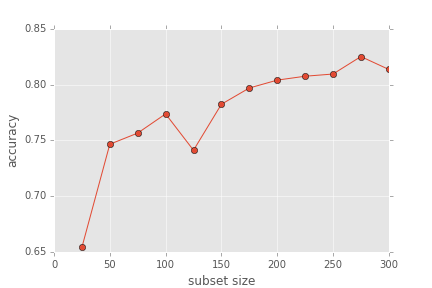}
\caption{\label{fig:acc_by_samplesize} Average accuracy of RF on subsets of isolates of increasing subset size. \m{k=10}.
}
\vspace{-12pt}
\end{figure}

\begin{figure}[!ht]
\centering
\includegraphics[width=0.4 \textwidth]{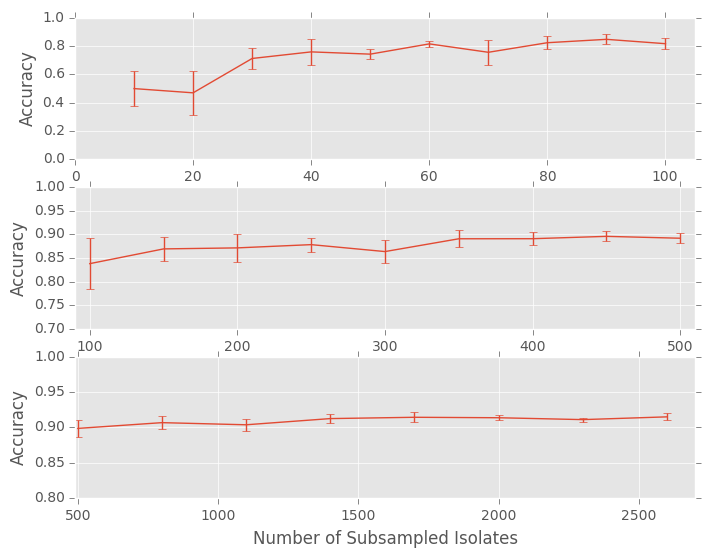}
\caption{Accuracy of Random Forest averaged over various sample sizes. \m{k=10}.\label{fig:figure_1}}
\vspace{-12pt}
\end{figure}

RF performs remarkably well in general. We predict AMR phenotypes using an \m{80/20} train/test split, obtaining accuracies as high as \m{92\%}. Figure \ref{fig:acc_by_samplesize} shows that the accuracies of the algorithm decrease as a function of the number of isolates trained on, while Figure \ref{fig:figure_1} shows that ultimately the accuracy plateaus on larger sample sizes. Despite lower accuracies, Figure \ref{fig:figure3} shows that the top ranked gene regions show stability even at a lower accuracy. Smaller datasets may show accuracies that are concerning, yet they may still be accurate enough to identify key locations of mutations that confer phenotype. Further tuning of the algorithm, preprocessing the data or feature selection would increase accuracy but would distract from our central thesis, which is chiefly that AML techniques can rapidly be applied to open and real world biological problems.

\subsection{Discussion}
We have transformed the problem of genotype to phenotype classification into a supervised machine learning task in order to exploit numerous algorithms and techniques available for classification. We believe the above demonstrates that the performance of AML techniques can be suitably used to investigate biological functions involved in AMR and present some opportunities for such analysis below. These corespond to differing objectives which we list below:

\textbf{Objective I, Maximize classification accuracy:}
Maximizing classification accuracy is incredibly important for long-term deep investigations of genotypic changes in relationship to phenotype. This approach offers great value when considering stable mutations and would allow for the identification of a genotypic fingerprint (i.e a unique genetic string indicative of a phenotype) of a single type of AMR. This insight could be used for construction of a biological assay (a devices that provides rapid diagnosis in the field). A large collection of genetically similar isolates undergoing a similar mutation conferring resistance would be required, but a nuanced understanding of what defines resistance would be achieved. This approach is perhaps most similar to the "leaderboard" competitions for AML. \emph{Example use}: 
design of an extremely accurate test for wide spread genetically similar outbreaks.

\textbf{Objective II, Maximize generalization accuracy:}
Generalization of classifiers for AMR phenotypes can be managed across different bacteria for the same antibiotic or across antibiotics for the same bacteria. It is known that antibiotics affect common genes and gene functions in bacteria, yet little is known about the generalization of the AMR genotype across different bacteria.
\emph{Example use}: 
designing a general purpose screening for AMR genotypes in environmental samples or designing an algorithmic tool for prioritizing first-line antibiotics in a new outbreak.

\textbf{Objective III, Aggregate feature selection:}
Feature selection provides crucial insight into biological processes that confer AMR. The most productive way to cluster features remains an open problem and one that could yield important insight for biologists. In our representation of the problem features are highly redundant and collapsing these redundant features into meaningful clusters, or blocks, provides powerful indication of the region of mutation that confers AMR.
\emph{Example use}: a small fast moving outbreak requires a corse analysis to identify regions of mutation in order to choose an appropriate antibiotic.

\vspace{-10pt}
\section{Conclusion}
We are far from the first to identify computational biology as a productive area of research for AML researchers. There is a concerted effort in the biostatistical community to perform outreach towards biologists. While the separation in data analysis skills and domain knowlege between a typical biologists and a typical statistician is very wide, we believe the AML community and in particular the Data-for-Good community have both the technical and dispositional skills to manage such a divide. Furthermore, we believe that the influx of aggregated datasets will provide both opportunity and need for such professionals.


\bibliography{Bib2}
\bibliographystyle{icml2016}

\end{document}